\documentclass[letterpaper, 10 pt, conference]{ieeeconf}  
\IEEEoverridecommandlockouts                              
\usepackage{graphicx} 
\usepackage{amsmath} 
\usepackage{amssymb}  
\graphicspath{{figures/}}

\usepackage{cite}
\usepackage{booktabs} 
\usepackage{siunitx} 
\usepackage{cleveref}

\usepackage{optidef}
\usepackage[dvipsnames]{xcolor}
\usepackage{mathtools}
\usepackage{subfig}
\usepackage[ruled]{algorithm2e}
\SetKwInput{kwInit}{Init}
\SetKwComment{Comment}{$\triangleright$\ }{}
\SetCommentSty{itshape}

\usepackage[]{todonotes}

\usepackage{etoolbox}

\title{\LARGE \bf
Natural Gradient Shared Control}

\author{Yoojin Oh$^{1}$, Shao-Wen Wu$^{1}$, Marc Toussaint$^{1,2}$ and Jim Mainprice$^{1,3}$\\
\vspace{0.1cm}
\authorblockA{\tt{\small{firstname.lastname@ipvs.uni-stuttgart.de}}}
\authorblockA{$^1$Machine Learning and Robotics Lab, University of Stuttgart, Germany}
\authorblockA{$^2$Learning and Intelligent Systems Lab ;  TU Berlin ; Berlin, Germany}
\authorblockA{$^3$Max Planck Institute for Intelligent Systems ;  IS-MPI ; T{\"u}bingen, Germany}
\vspace{-0.8cm}
}

\begin{document}
\bstctlcite{IEEEexample:BSTcontrol}

\maketitle
\thispagestyle{empty}
\pagestyle{empty}

\begin{abstract}

We propose a formalism for shared control,
which is the problem of defining a policy that blends user control
and autonomous control.
The challenge posed by the shared autonomy system is to maintain
user control authority while allowing the robot to support
the user. This can be done by enforcing constraints or acting optimally when the intent is clear.
Our proposed solution relies on
natural gradients emerging from the divergence constraint
between the robot and the shared policy.
We approximate the Fisher information 
by sampling a learned robot policy and computing the local gradient
to augment the user control when necessary.
A user study performed on a manipulation task demonstrates that our approach allows for more efficient task completion while keeping control authority
against a number of baseline methods.
\end{abstract}

\section{INTRODUCTION}
Intelligent robots can substitute or assist humans to accomplish complicated and laborious tasks. 
They are becoming present in our lives from production lines to hospitals,
and our daily homes. However, many applications remain challenging for
robots to function in full automation especially in dynamic environments (e.g. driving in extreme
weather conditions, disaster recovery) and robots still require human intervention.
Hence, we emphasize the need to study how to effectively balance human control and autonomy.

\textit{Shared control} has been studied 
to exploit the maximum performance of a robot system 
by combining human understanding and decision making with robot computation and execution capabilities. 
A linear blending paradigm introduced by Dragan et. al~\cite{dragan2013policy} is still widely applied 
in many shared control frameworks~\cite{goil2013using,anderson2014experimental,gao2014contextual}. 
In the approach, the amount of arbitration is dependent on the confidence of user prediction.
When the robot predicts the user's intent with high confidence,
the user often loses control authority.
This has been reported to generate mixed preferences from users
where some users prefer to keep control authority despite longer completion times~\cite{kim2011autonomy, Javdani:2018bt}.
Additionally, when assistance is against the user's intentions
this approach can aggravate the user's workload~\cite{dragan2013policy};
the user ``fights" against the assistance rather than gain help from it.

Some works have taken the option to
allocate maximum control authority to the user
by minimally applying assistance only when it is necessary.
Broad et. al.~\cite{broad2018operation,broad2019highly} introduced
minimum intervention shared control that computes whether the control signal 
leads to an unsafe state and replaces the user control if so.
However, these works usually model the problem by selecting
an action within a feasibility set. 

\begin{figure}[t]
\vspace{-.4cm}
  \includegraphics[clip,trim={0cm 2.7cm 0cm 2cm},width=1\linewidth]{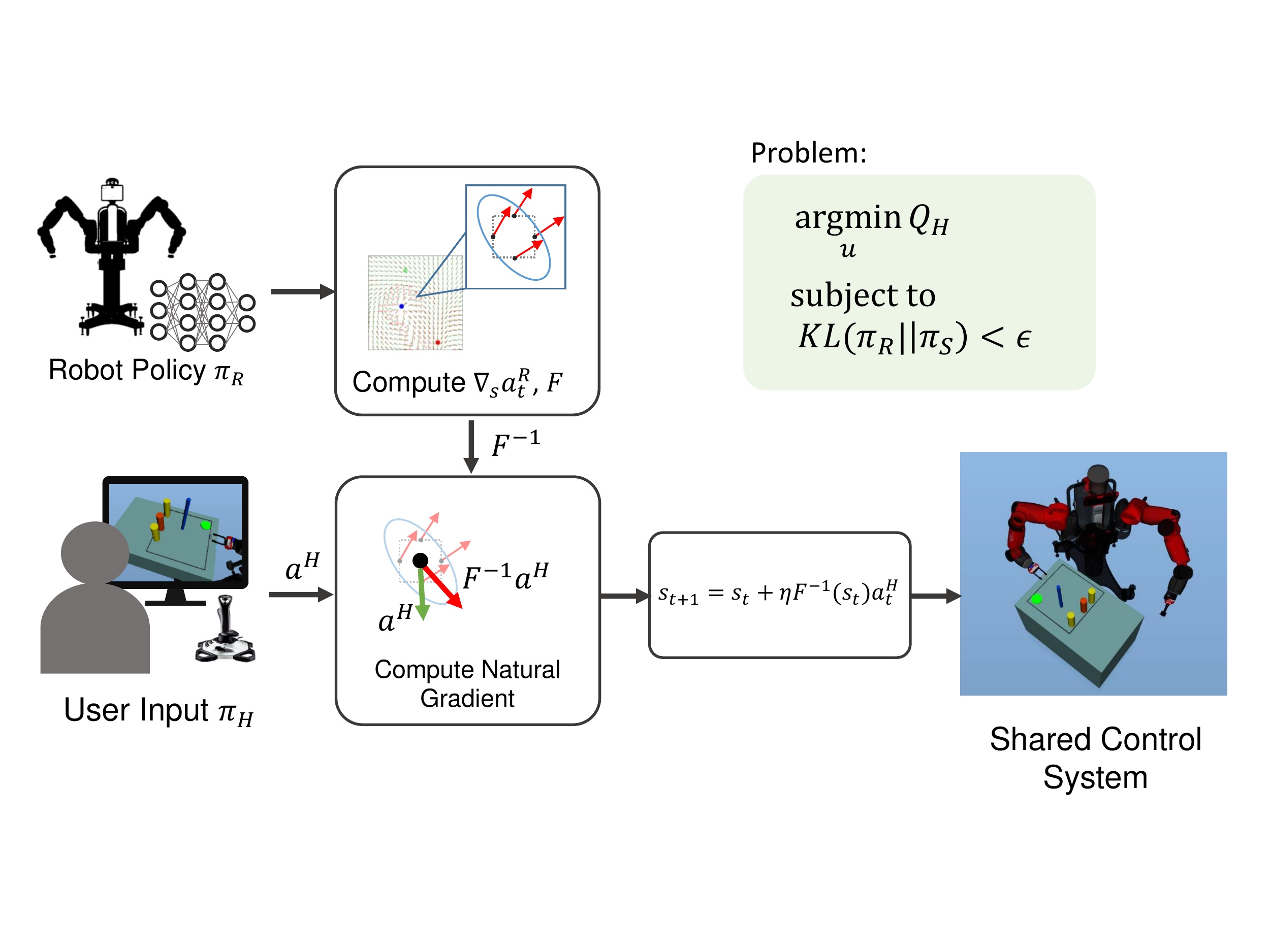}
  \vspace{-.7cm}
  \caption{
  		Overview of the shared control system.
  		We approximate the Fisher information matrix $F$ by sampling the robot policy and computing a local gradient. $F^{-1}$ augments the user policy, resulting in a natural gradient step update.}
  \label{fig:workflow}
 \vspace{-.5cm}
\end{figure}

In this work, we formulate shared control as an optimization
problem, as shown in Figure~\ref{fig:workflow}. The shared control action is chosen to maximize
the user's internal action-value function while we constrain the
shared control policy not to diverge away from the autonomous robot policy.
We construct the Fisher information matrix which expresses 
how sensitive a distribution changes in the local neighborhood
of the state. 
When the robot policy is represented 
as a vector field over the state space, the user can maintain more control authority
in regions where the field does not diverge.
On the contrary, the inverse Fisher information matrix adjusts the user's actions so that the robot gains more authority where the robot policy is rapidly changing in the local region (e.g. near an obstacle or near a goal). 
Utilizing the Fisher information matrix we introduce the term ``Natural gradient shared control".


To assess the efficacy of the approach, we define a teleoperation task 
where a user performs pick-and-place tasks with a simulated robot arm and compare the quantitative metrics. 
We show that our shared control paradigm can assist the user towards accomplishing the goal while enabling more control authority to the user.

We summarize our main contributions as the following:
\begin{itemize}
	\item A shared control paradigm that relies on natural gradients emerging from the divergence constraint between the robot and the shared policy
	\item Estimating the Fisher information matrix by sampling the autonomous robot policy and computing the local gradient using the finite difference method
	\item Generating an autonomous robot policy that only relies on task space states and can be regressed over the entire state space 
	\item Quantitative results of a preliminary human user study including 16 participants that shows the potency of the proposed paradigm to allow more control authority to the user
\end{itemize}

This paper is structured as follows: we present relevant related work in Section~\ref{sec:rel_work}. We review the background theory that underlies our concept of using natural gradients in Section~\ref{sec:natural_gradients}.  Sections~\ref{sec:NGSC}~and~\ref{sec:robot_policy} introduce our framework and explain the implementation. The overview of our pilot user study method is described in Section~\ref{sec:experiments} and we present the results in Section~\ref{sec:results}. Finally conclusions are drawn in Section~\ref{sec:conclusions}.

\section{RELATED WORK}
\label{sec:rel_work}
\subsection{Shared Control}
Shared control refers to a cooperation between the user and the autonomous robot agent to achieve a common task at a control level~\cite{flemisch2016shared}. 
In previous works, different paradigms for shared control have been developed depending on the task and purpose of the application. We categorize the paradigms in three groups depending on how the agents share control: control switching, direct blending, and indirect blending.

\textbf{Control switching}
We refer to control switching as allocating all-or-none assistance during control. It is a discrete switching between full autonomy and direct control depending on a predefined circumstance. Control switching can be momentary: at each state the robot evaluates whether to take over depending on the confidence of the user's intentions~\cite{dragan2013policy}~(aggressive mode); or to prevent the system from entering an unsafe state~\cite{broad2018operation}. 
Control switching is also referred to as \textit{Traded control} when the robot takes over and autonomously executes a sub-task over a sequence of timesteps~\cite{kofman2005teleoperation,smith2008teleoperation, phillips2016autonomy}. 

\textbf{Direct blending} It involves explicitly combining both agents' control using an arbitration function. The above \textit{control switching} corresponds to when the step function is used as the arbitration function.  
Approaches include using a linear function~\cite{dragan2013policy,anderson2014experimental,gao2014contextual}, a sigmoid function~\cite{muelling2015autonomy}, specifically tuned~\cite{gopinath2016human,jain2016approach}, or learned~\cite{goil2013using,oh2019learning}. 
It can be tedious to define and tune an arbitration function that generalizes between users and tasks. As a result, the blended action may be worse than following one of either policy. 

\textbf{Indirect blending}
We relate to \textit{Indirect blending} when the shared control is a result of an optimization problem. Javdani et al.~\cite{Javdani:2018bt} formulated the problem as a POMDP and approximated using hindsight optimization to provide assistance which minimizes the expected cost-to-go for an unknown goal. Reddy et al.~\cite{reddy2018shared} used deep Q-learning to select an action closest to the user's suggestion while being suboptimal in discrete states. Broad et al.~\cite{broad2019highly} introduced shared control with ``minimal intervention", such that the human's command is perturbed when the system leads to an inevitable collision state. Similarly, our work does not directly blend the controls despite the presence of the autonomous robot action but rather computes a metric based on the topology of the autonomous policy in the local neighborhood and adjusts the user commands. 

\subsection{Natural Gradients}
The natural gradient adjusts the direction of the standard gradient according to the underlying Riemannian structure of the parameter space. It is effective when the direction of the standard gradient descent does not represent the steepest descent direction of the cost function in the parameter space which lead to poor convergence~\cite{amari1998natural,amari1998why}. 

The natural gradient is advantageous as it is invariant under coordinate transformations and unlike Newton's method, it doesn't assume the cost function as locally-quadratic. It is widely applicable to non-linear functions in neural networks~\cite{amari1998natural}, including reinforcement learning methods such as policy gradient~\cite{kakade2002natural} and actor-critic~\cite{peters2008natural}.
In Schulman et. al~\cite{schulman2015trust}, a trust-region is defined by constraining the KL-divergence between the old and the new policy during policy updates. Similarly, our shared control framework imposes a constraint on the KL-divergence of the autonomous policy and the shared control policy such that the shared control policy does not diverge far from the autonomous robot policy.

\section{Understanding Natural Gradients}
\label{sec:natural_gradients}
Gradient descent is a procedure to locally minimize a function $f(\theta)$ by updating its parameters. It can be written as an optimization problem where we minimize a linear approximation of the function $f(\theta)$ subject to the constraint that the distance in the step $\delta\theta = \theta-\theta_t$ is marginal~\cite{ratliff2013information}:
\begin{argmini}
	{\theta}{f(\theta_t)+\nabla f(\theta_t)^T(\theta-\theta_t)}
	{}{}
	\label{opt:opt_problem}
	\addConstraint{||\theta-\theta_t||^2_G = \epsilon^2}
\end{argmini}
The natural gradient descent computes the distance over the manifold in the Riemannian space that the coordinates parameterize. It is measured as $|\delta\theta|^2\equiv \sum_{ij}G_{ij}(\theta)\delta\theta_i \delta\theta_j=\delta\mathbf{\theta}^TG(\theta)\delta\mathbf{\theta}$ (in vector notation) where $G(\theta)$ is called the \textit{Riemannian metric tensor} which characterizes the intrinsic curvature of a particular manifold in $N$-dimensional space. The natural gradient is then defined as 
\begin{equation}
	\tilde{\nabla}f=F^{-1}\nabla f
	\label{eqn:natural_gradient}
\end{equation}
It is equivalent to standard gradient descent when $G(\theta)$ is the identity matrix, such that $|\delta\theta|^2$ is equivalent to the squared Euclidean norm~\cite{amari1998natural}. When $G(\theta)$ is a positive-definite Hessian, it corresponds to taking Newton direction steps in the Newton's method. However, Newton's method and natural gradient adaptation differ in general: $G(\theta)$ is always positive-definite by construction and the Hessian may not be~\cite{amari1998why}. When $G(\theta)$ lies in the parameter space of a statistical model, $G(\theta)$ refers to the Fisher information matrix $F(\theta)$. 

The Fisher information $F(\theta)$, by definition, measures the expectation of the overall sensitivity of a probability distribution $p(x|\theta)$ to changes of $\theta$. It is defined as the variance of the $\textit{score}=\frac{d}{d\theta}\log p(x|\theta)$, which indicates the sensitivity of the model to changes in $\theta$ \cite{ly2017tutorial}. 
\begin{equation}
	\label{eqn:fisher_def}
	\begin{aligned}
	F(\theta)
	&=\displaystyle \mathop{\mathbb{E}}_{p(x|\theta)}\Big[ \Big(\frac{d}{d\theta}\log p(x|\theta)\Big)^2\Big]  \\
	&= -\int p(x|\theta)\frac{d^2}{d\theta^2}\log p(x|\theta)dx \footnotemark\\
	\end{aligned}
\end{equation} 
\footnotetext{It is equivalently defined under mild regularity conditions~\cite{ly2017tutorial}} 
Most importantly to our interest, the Fisher information is the second order derivative of the KL-divergence \cite{ratliff2013information}\cite{Kristiadi2020blog}. 
\begin{equation}
	\begin{aligned}
	F(\theta) 
	&= -\int p(x|\theta)\nabla^2_{\theta'}\log p(x|\theta')\big|_{\theta'=\theta}dx  \\
	&= H_{\mathrm{KL}(p(x|\theta ||p(x|\theta')))}  \\
	\end{aligned}
	\label{eqn:fisher_kl}
\end{equation}
This gives the key to the connection between natural gradient and the KL-divergence, where KL-divergence is the function to measure the ``distance\footnote{KL-divergence is not formally a distance metric since it is not symmetric}" in gradient descent~\cite{martens2014new} as below:
\begin{equation}
	\mathrm{KL}(p(x|\theta) || p(x|\theta')) 
	\approx \frac{1}{2}\delta\theta^T F(\theta) \delta\theta
	\label{eqn:kl_approx}
\end{equation}
We use Equations~\ref{eqn:fisher_def}-\ref{eqn:kl_approx} in the following section to solve our optimization problem and approximate the Fisher information matrix.

\section{Natural Gradient Shared Control}
\label{sec:NGSC}
\subsection{Expressing Shared Control as an Optimization Problem}
Let $s \in \mathcal{S}$ be the state of the system. Let $a^H \in \mathcal{A^H}$ as the user action, $a^R \in \mathcal{A^R}$ be the autonomous robot action, and $u \in \mathcal{U}$ be the shared control action. The human and the robot agent each select actions following their stochastic policies, $\pi_H$, and $\pi_R$.
Our goal is to find a shared control policy $\pi_S$ 
that solves the following optimization problem.
\begin{argmaxi}
	{u_t}{Q_H (s_t, u_t)}
	{}{}
	\label{opt:problem}
	\addConstraint{\mathrm{KL}(\pi_R \| \pi_S) < \epsilon}
\end{argmaxi}

The shared control policy is chosen to maximize the user's internal action-value function $Q_H(s_t, u_t)$ at each step. 
Although it is possible to learn $Q_H$ using methods such as MaxEnt IOC~\cite{ziebart2009planning}, predicting the user controls $a^H_t$ can be challenging due to interpersonal differences. Instead, we regard the user action as an estimate of $\nabla_s Q_H(s_t, a^H_t)$ at each step.

The constraint on the KL-divergence between the robot policy and the shared policy ensures that the shared policy does not deviate far from the autonomous robot policy.
The problem can be expressed using a Lagrange Multiplier,
assuming a linear approximation of our objective $\nabla_s Q_H(s_t, a^H_t)$ and a quadratic approximation of
the KL-divergence constraint. 
Solving this approximation of the Lagrangian leads
to an update rule which introduces natural gradient adaptation.
\begin{algorithm}[t]
\DontPrintSemicolon
\SetAlgoNoLine
 \For{$t=1,T$}{
  Observe user action $a^H_t$\;
  \ForEach{$g \in \mathcal{G}$}{
    Compute belief $b_{t,g}$\;
    \SetKwFunction{FMain}{ComputeFisher}
    $F_g(s_t)=$ \FMain{$s_t, g$}\;
  }
  $F(s_t)^{-1} = \sum_g b_{t,g}F_{t,g}^{-1}(s_t)$ \Comment*[r]{weighted sum over goals}
  $u_t \leftarrow F(s_t)^{-1}a^H_t$\Comment*[r]{compute shared action} 
  $s \leftarrow s + \eta u_t$ \Comment*[r]{update state}
 }
 \caption{Natural Gradient Shared Control}
 \label{alg:NGSC} 
 \vspace{-.1cm}
\end{algorithm}

\subsection{Natural Gradient Shared Control}
We introduce a new control paradigm for shared control using natural gradient adaptation. Note that our goal is to find the action that maximizes $Q_H$, resulting in taking gradient steps in the direction of ascent.
\begin{equation}
	\label{eqn:state_update}
	\begin{aligned}
	s_{t+1} &= s_t + \eta F(s_t)^{-1} \nabla_s Q_H(s_t, a^H_t)  \\
	&=s_t + \eta u_t
	\end{aligned}
\end{equation} 
$\eta$ is the step size and the natural gradient in Equation~\ref{eqn:natural_gradient} corresponds to the shared control action ${u_t \sim\pi_S(\cdot|s_t, a^H_t, a^R_t)}$. We utilize the approximation ${a^H_t\propto\nabla_sQ_H(s_t,a^H_t)}$ in Equation~\ref{eqn:shared_action}. The proportionality constant is absorbed by the step size $\eta$. The overall algorithm is summarized in Algorithm~\ref{alg:NGSC}.
\begin{align}	
	u_t &= F(s_t)^{-1}\nabla_s Q_H(s_t, a^H_t) \\
	&=F(s_t)^{-1}a^H_t   \label{eqn:shared_action}
\end{align}

The Fisher information matrix $F(s_t)$ can be interpreted as the sensitivity of the autonomous robot policy $\pi_R$ to changes of the parameter.

Intuitively, a deterministic robot policy regressed over the
whole state space defines a vector field.
This vector field integrates information about which
optimality and constraint trade-offs are made about the underlying actions.
For example when an obstacle is in an environment, it acts as a \textit{source} (positive divergence) in the vector field
resulting in a repulsive action.
When the policy is goal-directed, the goal acts as a \textit{sink} (negative divergence). The vectors around the goal point inward.
$F$ measures how sensitive the field changes and emphasizes or discounts towards certain directions of $u_t$. 

\begin{algorithm}[t]
\DontPrintSemicolon
\SetAlgoNoLine
 \kwInit{
 Load pre-trained robot policy model $f_{nn}$ \; 
 }
 \KwIn{$s_t, g$}
 \KwOut{$F_{t,g}$}
 \SetKwFunction{FMain}{ComputeFisher}
 \SetKwProg{Fn}{Function}{:}{}
 \Fn{\FMain{$s_t, g$}}{
 	Sample set of states $\mathcal{S}=\{(s_i,g)\}^N_{i=1}$\;
    Infer robot actions $\mathcal{D} = f_{nn}(\mathcal{S})$\;
    Fit LWR model $L_g=\textrm{LWR}(\mathcal{S},\mathcal{D})$\;
    $\tilde{a}^R_{t,g} \leftarrow L_g(s_t)$ \Comment*[r]{query action from LWR}
    $H_{t,g} \leftarrow \nabla_s \tilde{a}^R_{t,g}$ \Comment*[r]{compute Jacobian of action}
    $F_{t,g} \leftarrow \frac{1}{2} (H_{t,g}+H_{t,g}^T)$ \Comment*[r]{ensure symmetry}
    \textbf{return} $F_{t,g}$\;
 }
 \textbf{End Function}\;
 \caption{Computing Fisher Information Matrix}
 \label{alg:fisher}
\end{algorithm}

\subsection{Computing the Fisher Information Matrix}
\label{ssec:compute_fisher}
We approximate $F$ as the curvature of the robot's action-value function at a given state:
\begin{align}
\label{eqn:fisher_definition}
	F(s_t) &= \displaystyle \mathop{\mathbb{E}}_{\pi_H}
		[\nabla_s\log \pi_R(a^R_t|s_t)\nabla_s\log \pi_R(a^R_t|s_t)^T]  \\
	&\approx \nabla_s^2 Q_R(s_t, a_t)
\end{align}

Keeping in mind the connection between $F$ and the KL-divergence
and the asymmetry of the KL-divergence,
one may ask how dependent is $F$
to the underlying assumption of $\mathrm{KL}(p || q)$ or $\mathrm{KL}(q||p)$.
It turns out that when $p$ and $q$ are close,
the KL-divergence is locally/asymptotically symmetric~\cite{martens2014new}.
Hence our definition for $F \approx \nabla_s^2 Q_R$ is equivalent
to integrating the user actions over all possible robot actions.

We describe our method for computing $F$ in Algorithm~\ref{alg:fisher}, where
$F$ is computed at each state for each goal.
We use Locally Weighted Regression (LWR) to fit a local model $L_g$ using a set of sampled states and actions inferred using a pre-trained model. 
A detailed explanation regarding LWR is described in Subsection~\ref{ssec:lwr}. As we consider action as an approximate of the first derivative of the Q-function, we consider the Jacobian of the robot action w.r.t. state $\nabla_s a^R_t$  as the Hessian of the Q-function. $\nabla_s \tilde{a}^R_t$ is the Jacobian computed using the finite difference method with actions $\tilde{a}^R_{t,g}$ from $L_g$. The Fisher information matrix $F$ is positive-definite by definition. However, the Jacobian computed using the finite difference method may not always be symmetric. Thus, We decompose the matrix as a sum of symmetric and a skew-symmetric matrix and apply the symmetric matrix. 

\begin{figure}[t]
\vspace{-0.7cm}
  \captionsetup[subfigure]{justification=centering}
  \centering
  \subfloat[Single goal]{\centering\includegraphics[clip,trim={3.3cm 1.0cm 3cm 1.0cm},width=0.32\linewidth]{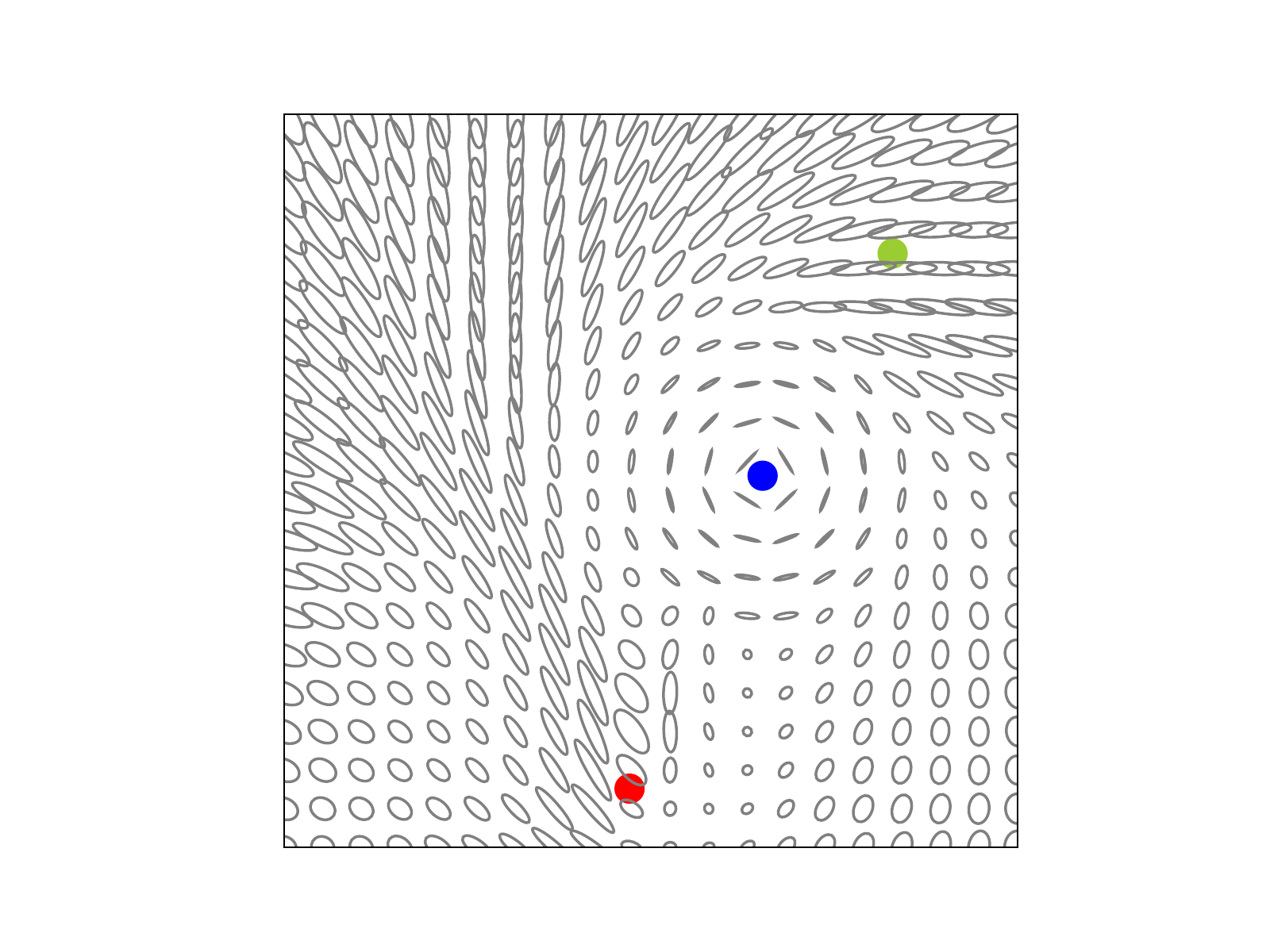}}
  \subfloat[Belief-weighted \newline over goals]{\centering\includegraphics[clip,trim={3.3cm 1.0cm 3cm 1.0cm},width=0.32\linewidth]{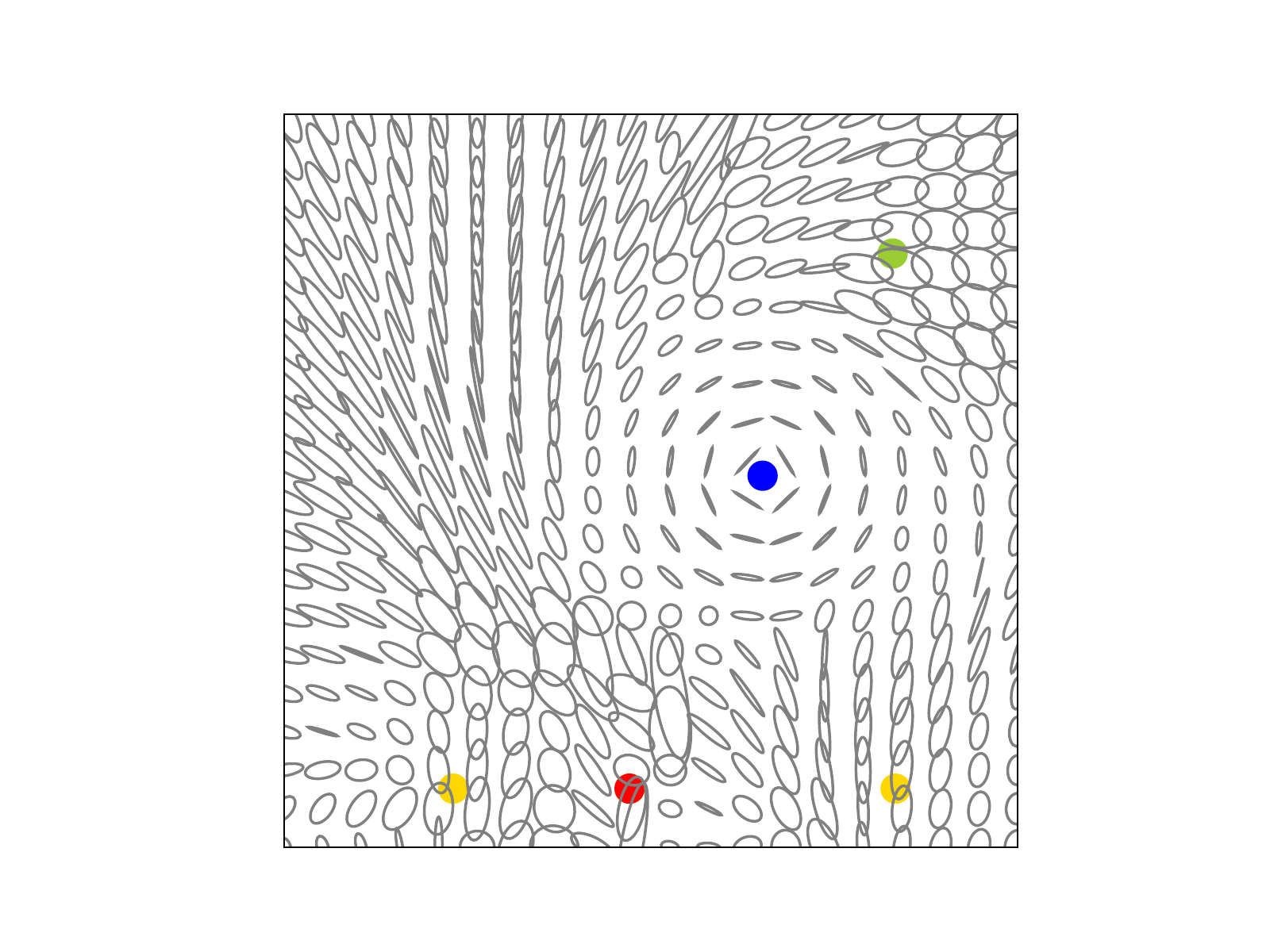}}
  \subfloat[Obstacle avoidance]{\centering\includegraphics[clip,trim={3.3cm 1.0cm 3cm 1.0cm},width=0.32\linewidth]{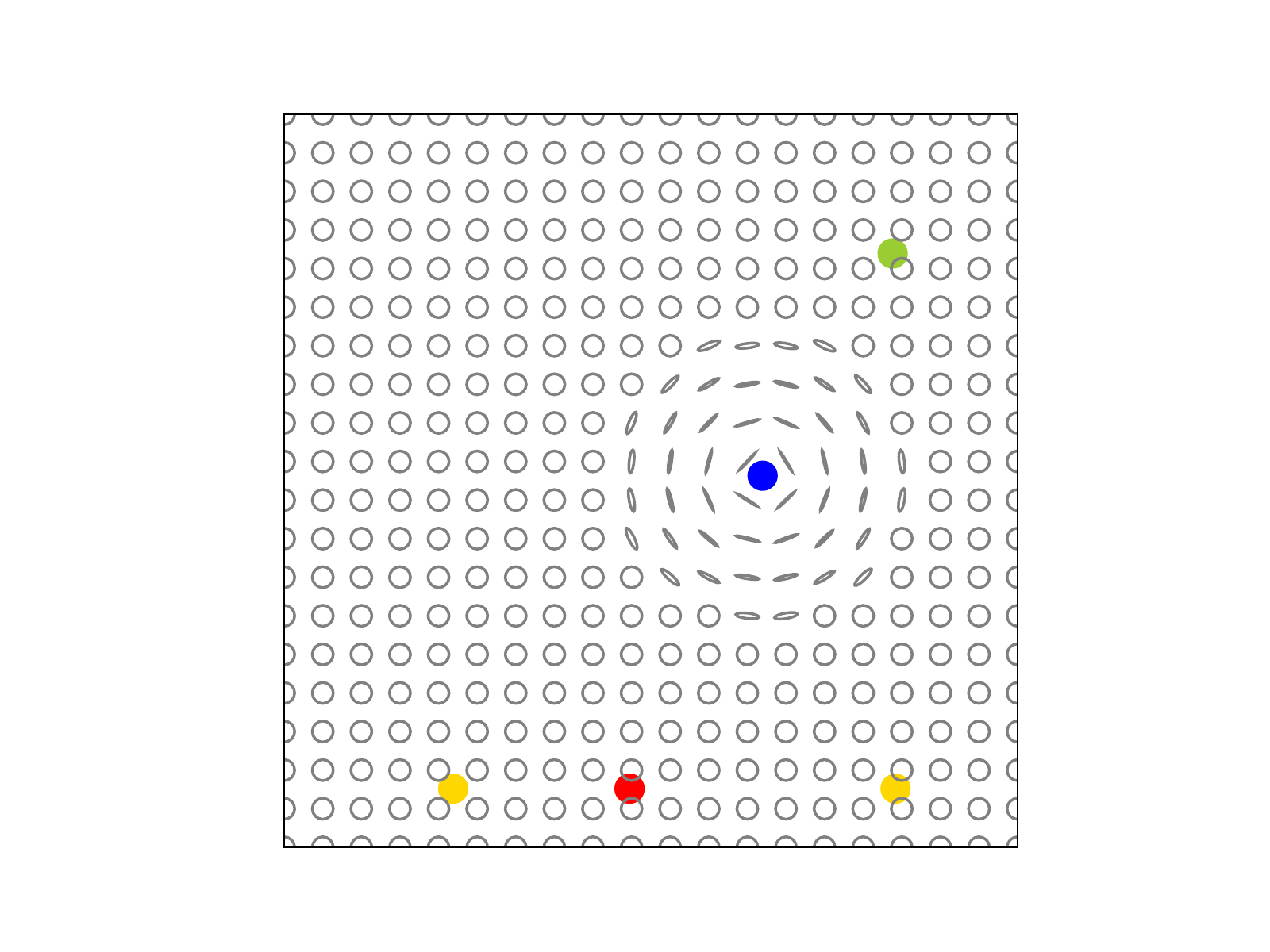}}
  \caption{Ellipse plots from the eigenvalues, eigenvectors of $F(s)^{-1}$ for (a) a single goal and (b) in the presence of multiple goals computed as weighted sum over beliefs, (c) obstacle avoidance}
  \label{fig:F_inv_ellipse}
  \vspace{-.7cm}
\end{figure}

The objective of the robot assistance can be flexibly defined depending on the cost function that the robot policy optimizes. Figure~\ref{fig:F_inv_ellipse} shows the $F(s)^{-1}$ computed over the state space with different assistance modes; the robot assistance can be goal-directed (in Figure~\ref{fig:F_inv_ellipse}(a)) or it can minimally assist to avoid obstacles (in Figure~\ref{fig:F_inv_ellipse}(c)). The ellipse represents the direction that the user's action is stretched along. When the ellipse is close to a circle the user has more control authority over the system. When the ellipse is narrow, for example near an obstacle, the robot augments the user's action towards one direction.

In the case of goal-directed assistance with multiple goals, the robot must predict a goal among goal candidates. We compute $F(s_t)^{-1}$ as a weighted sum over the beliefs which represent the confidence of the goal prediction. Figure~\ref{fig:F_inv_ellipse}(b) shows $F(s_t)^{-1}$ computed using beliefs as a naive distance-based goal prediction. It is shown that when the confidence is low (right top corner, or above the left two objects), user gains more control authority.

\section{Autonomous Robot Policy Generation}
\label{sec:robot_policy}
\subsection{Learning a Robot Policy}
\label{ssec:learning_policy}
We train a neural network policy that imitates optimized robot trajectories. 
Given the system's state $s_t$ and a goal $g$, the network infers a normalized velocity $v_{gripper}$ and end-effector rotation $\mathbf{v}_{rot_Z}$ that represents an optimal action towards the goal.
The network is regressed over the whole state space, such that it can infer an optimal action at any state that the robot is located in. This is advantageous when the robot state is modified with user inputs.

\begin{figure}[t]
  \includegraphics[clip,trim={0cm 1.5cm 0cm 2.9cm},width=1\linewidth]{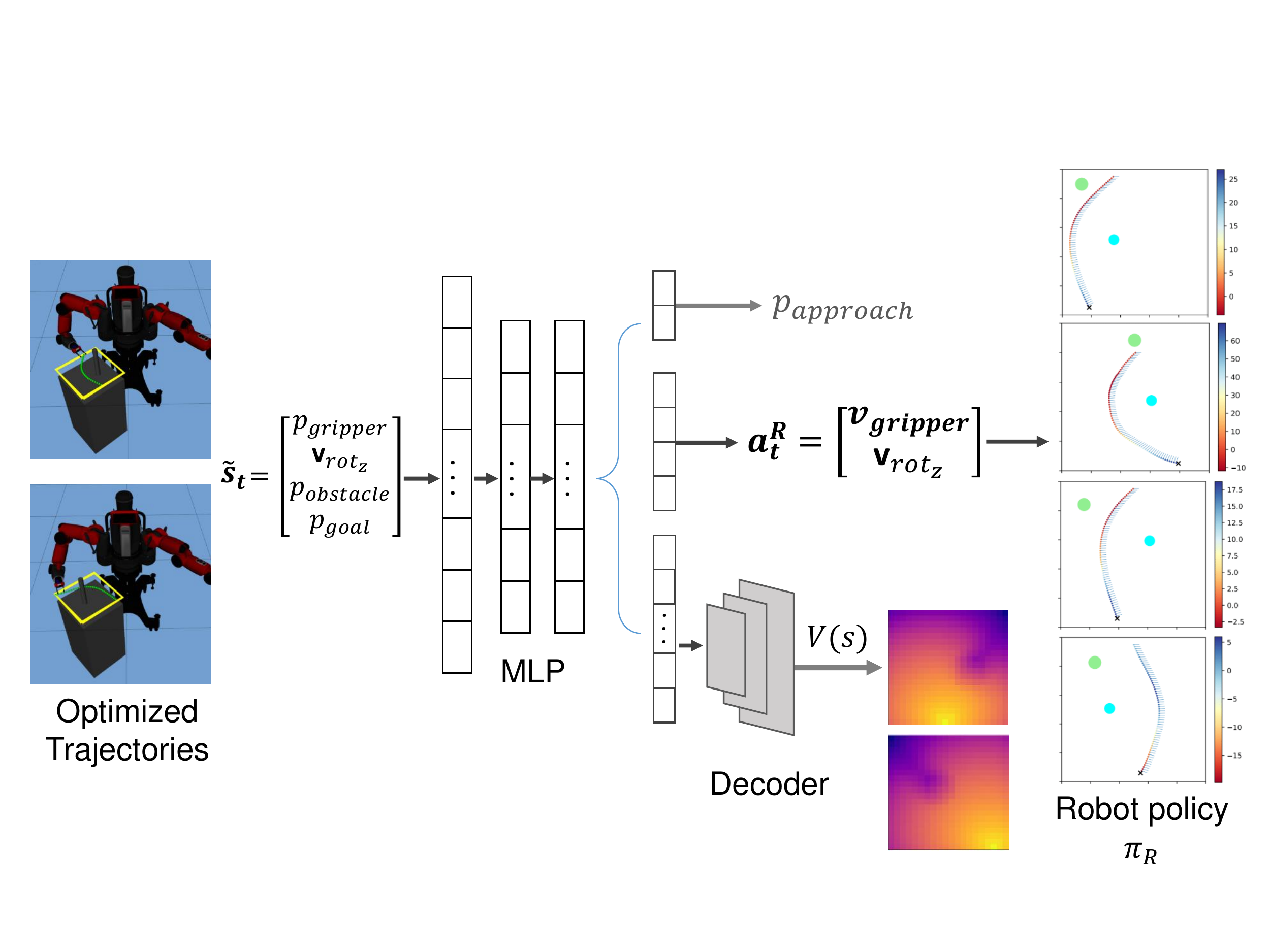}
  \caption{Overview of the training network. We use supervised learning to mimic optimal trajectory actions $a^R_t$ while using a multi-loss to infer additional features such as $p_{approach}$ and $V(s)$}
  \label{fig:network}
  \vspace{-.5cm}
\end{figure}

The overview of the training network is shown in Figure~\ref{fig:network}.
The state $\tilde{s}_t$ is a concatenation of robot and environment states consisting of: end-effector position $p_{gripper}$, vector components of current gripper rotation in z-axis 
\begin{equation}
\mathbf{v}_{rot_Z}=[\cos\varphi , \sin\varphi]^T
\end{equation} 
obstacle position $p_{obstacle}$, as well as the position $p_{goal}$ of the inferred goal. In the pick phase $p_{goal}$ refers to the position of the predicted goal object and in the place phase $p_{goal}$ refers to where the object should be placed. In our experiment, we constrict our environment to a 2-dim plane over the workspace.

The state and action space of our robot policy are solely defined in task space. 
Consequently, the network must learn to distinctively output end-effector actions 
such that it avoids collision when joint configurations are computed using inverse kinematics.
We collect training data by computing pick-and-place trajectories with random starting positions and environment configurations 
using a Gauss-Newton trajectory optimizer~\cite{mainprice2016warping} that is tuned to perform specific pick-and-place motions. 

To improve training, we use a multi-loss function to penalize/enforce specific behaviors. In the pick phase, we add an additional cost term to predict the approach position $p_{approach}$, which is a data point in the optimal trajectory that the gripper reaches when approaching the object. This ensures that the gripper approaches the object in the direction parallel to the gripper fingers for grasping. 

In addition, we learn a feature-based cost function of the environment and compute the \textit{state log partition function} $V(s)$ using MaxEnt IOC~\cite{ziebart2009planning,kitani2012activity}. The \textit{state log partition function} $V(s)$ is a soft estimate of the expected cost to reach the goal from a state $s$. Here, even though we limit the state/action space to 2-dim position and discrete neighboring actions, we were able to generate smoother trajectories by enforcing the network to reconstruct the learned $V(s)$ (shown as heatmaps in Figure~\ref{fig:workflow}) from the latent layer of the network using a series of convolutions and up-sampling layers. 
\begin{figure}[t]
  \centering
  \includegraphics[clip,trim={3cm 1cm 3cm 1cm}, width=.40\linewidth]{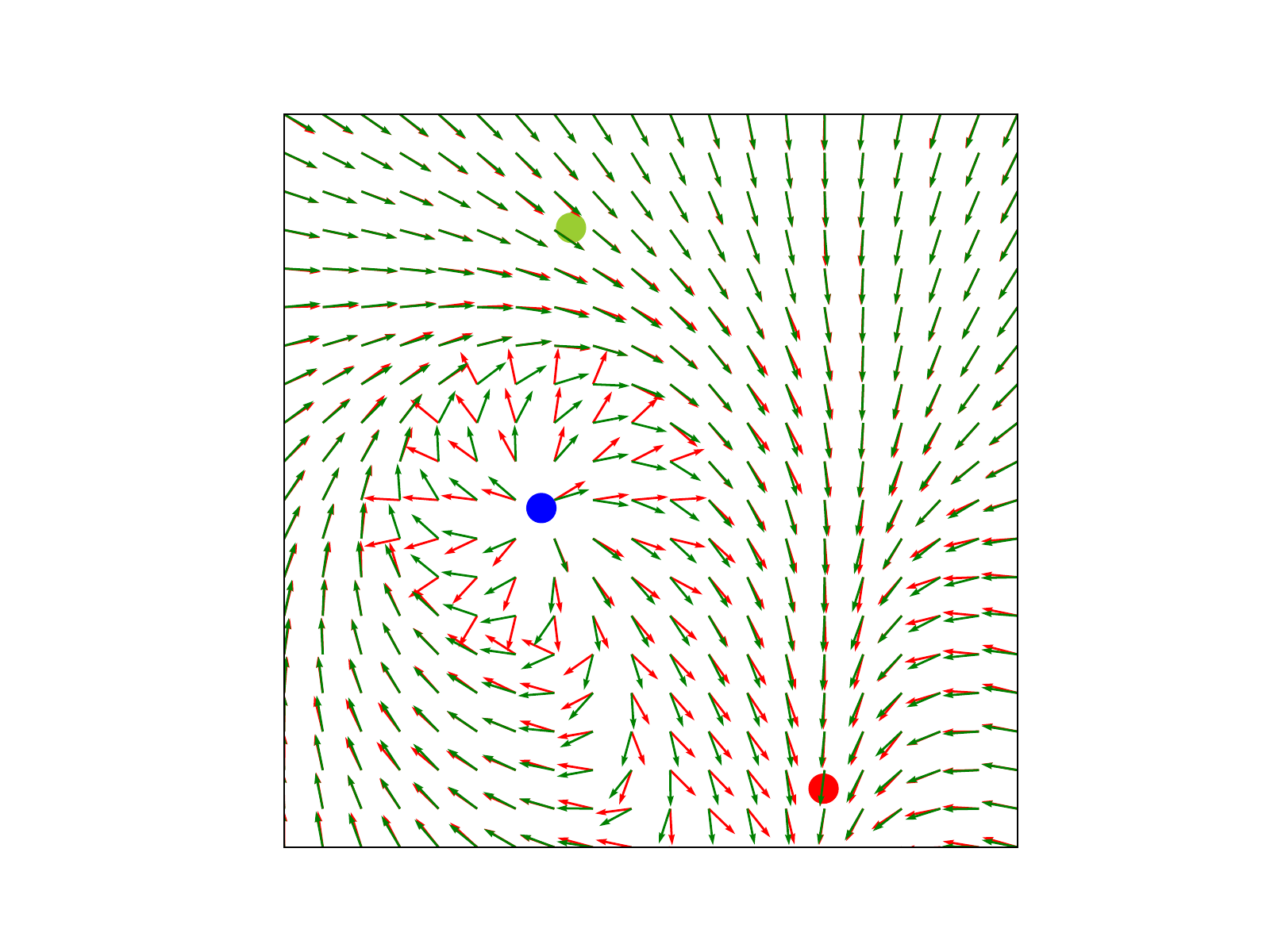}
  \centering
  \vspace{-.4cm}
  \includegraphics[clip,trim={3cm 1cm 3cm 1cm}, width=.40\linewidth]{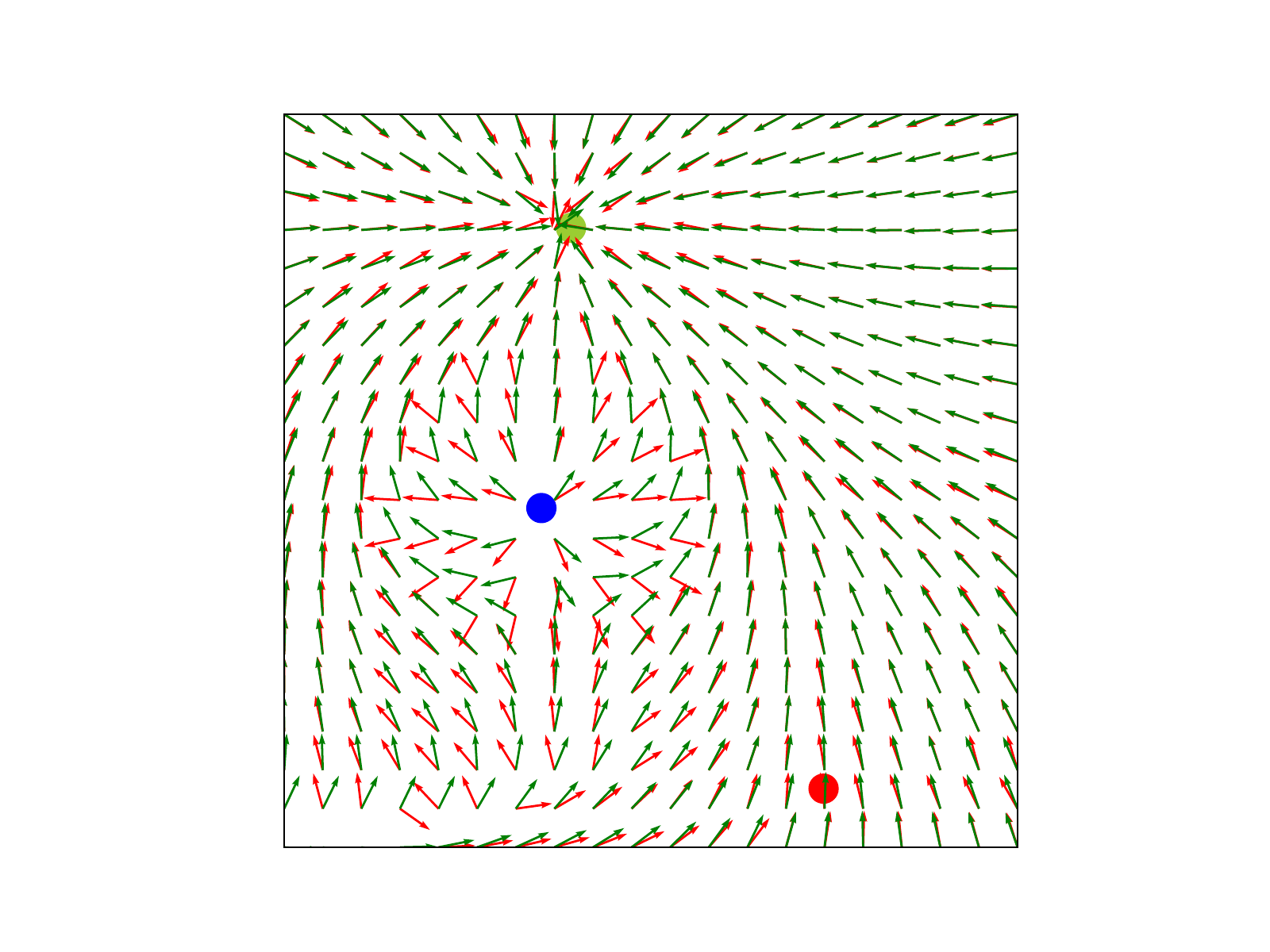}
  \subfloat[Pick policy]{\centering\includegraphics[clip,trim={3cm 1cm 3cm 1cm}, width=.40\linewidth]{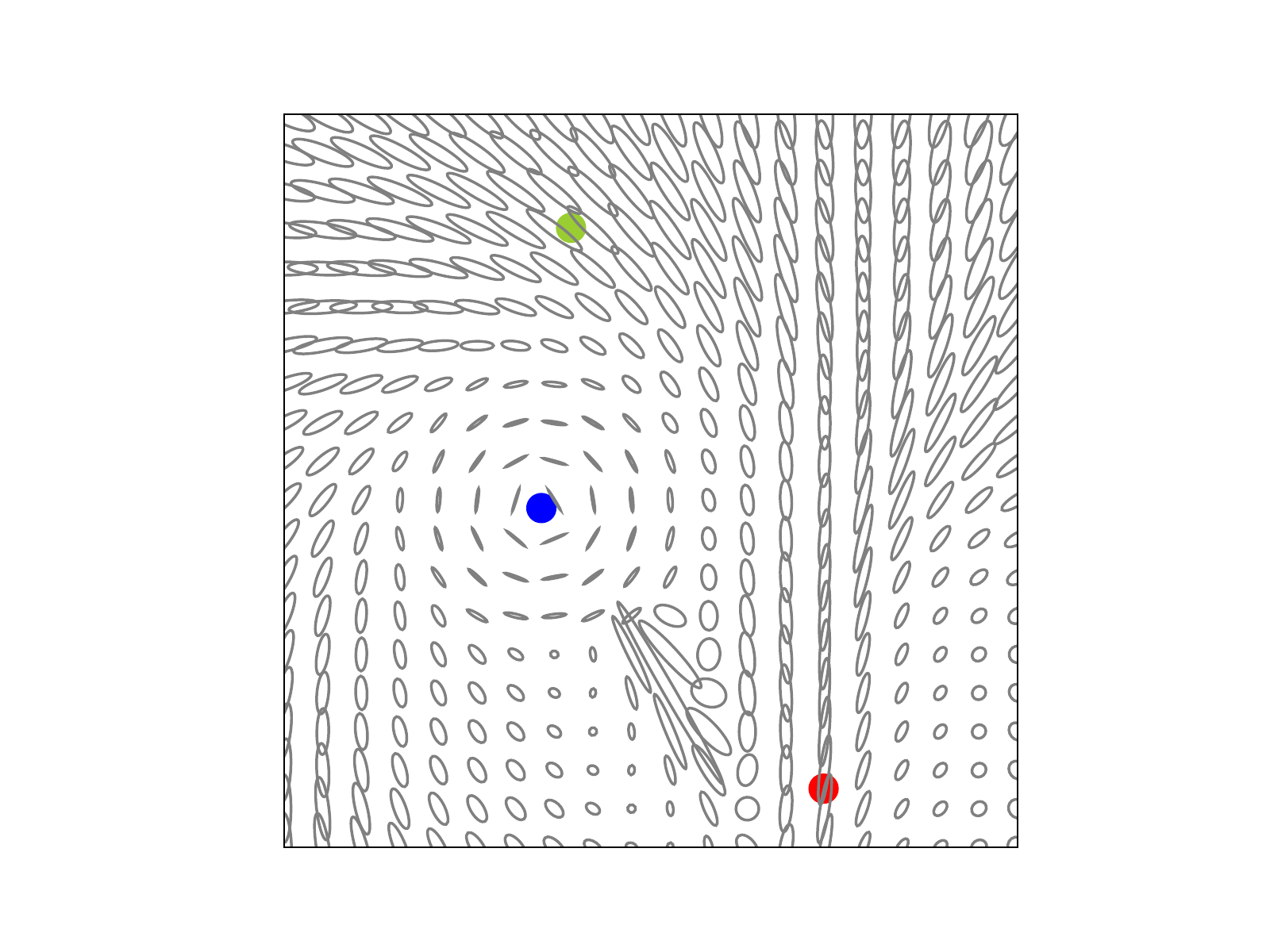}}
  \subfloat[Place policy]{\centering\includegraphics[clip,trim={3cm 1cm 3cm 1cm}, width=.40\linewidth]{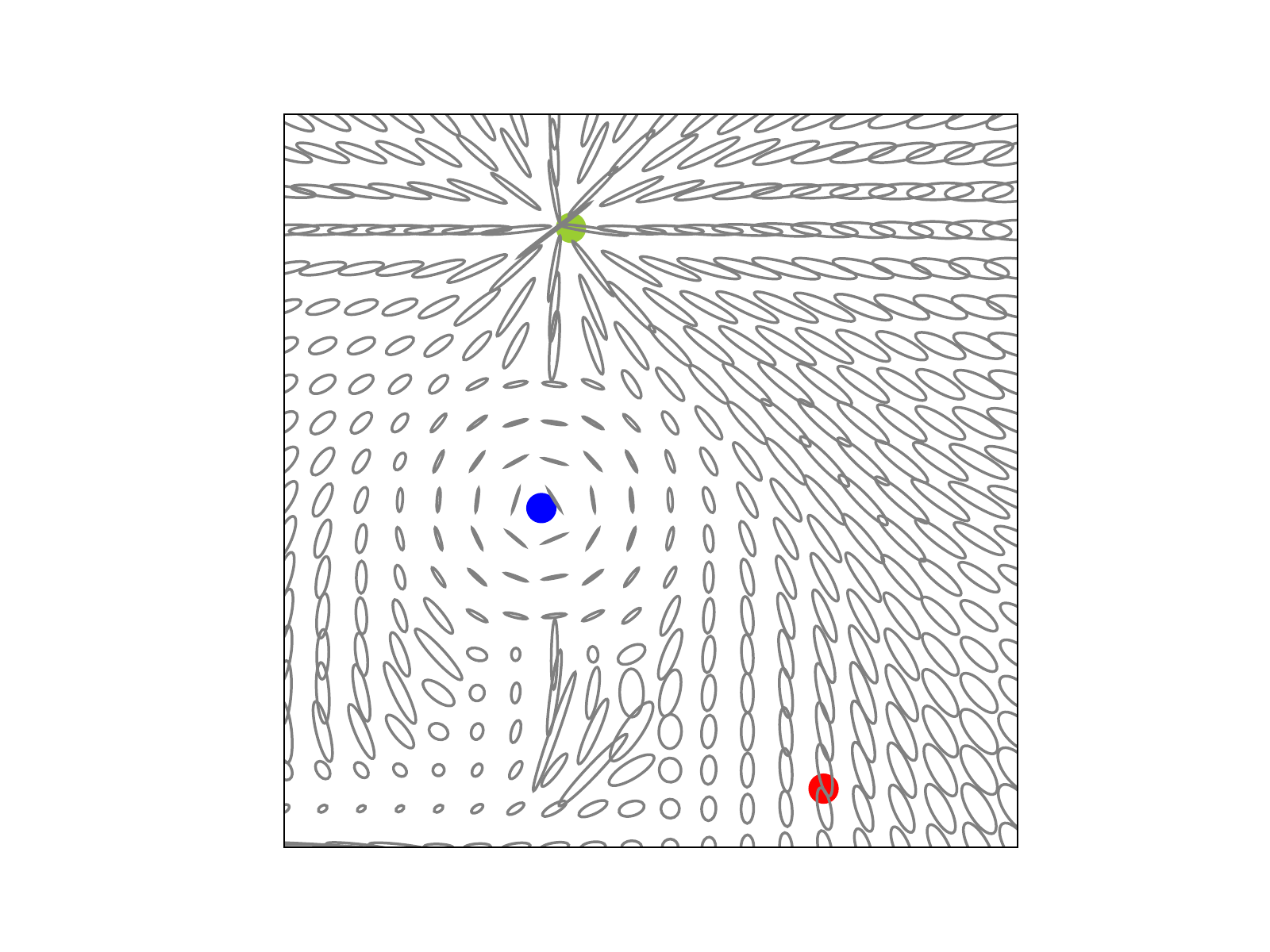}}
  \caption{First row: Vector field of the robot policy (green arrows) and the LWR fit policy (red arrows) projected onto position space. Second row: Ellipse plots from the eigenvalues and eigenvectors of $F(s)^{-1}$. Red, blue, green dots each correspond to positions of the object, obstacle, and goal.}
  \label{fig:robot_policy_lwr}
  \vspace{-.5cm}
\end{figure}

The robot policy is represented as a vector field over the state space, as shown in green arrows in the top row of  Figure~\ref{fig:robot_policy_lwr}. It shows positive divergence from the obstacle and negative divergence towards a goal.
When strictly following the inferred action at each state, the robot generates an autonomous pick-and-place trajectory with a success rate of 95\% in random environments.

\subsection{Fitting Locally Weighted Regression Models}
\label{ssec:lwr}
Locally Weighted Regression (LWR) fits a regression model that is valid only in the local region given a set of data points around the point of interest~\cite{atkeson1997locally}. 
Since the resulting model is linear, we fit the robot policy to a LWR model to reduce any non-linearity generated by the output of the network. The model is used to query data points to compute smoother Jacobians using the finite difference method, as part of the process to compute $F$ described in Subsection~\ref{ssec:compute_fisher}.

Another advantage of using LWR is that we can augment the model by providing additional data points since the model is fit only in the local region. This is useful to improve the behavior of the learned policy. For instance, we prefer the learned policy to enforce stronger obstacle avoidance, thus we provide additional data points computed using a signed distance field when fitting LWR near the obstacle.
 
The vector fields of the policies regressed using LWR are shown as red arrows in the top row in Figure~\ref{fig:robot_policy_lwr}. The LWR model successfully approximates the neural network policy, as well as augmenting a stronger repulsion in neighboring points of the obstacle.

\section{EXPERIMENTS}
\label{sec:experiments}
\subsection{Experimental System}
\begin{figure}[t]
  \includegraphics[width=.48\linewidth]{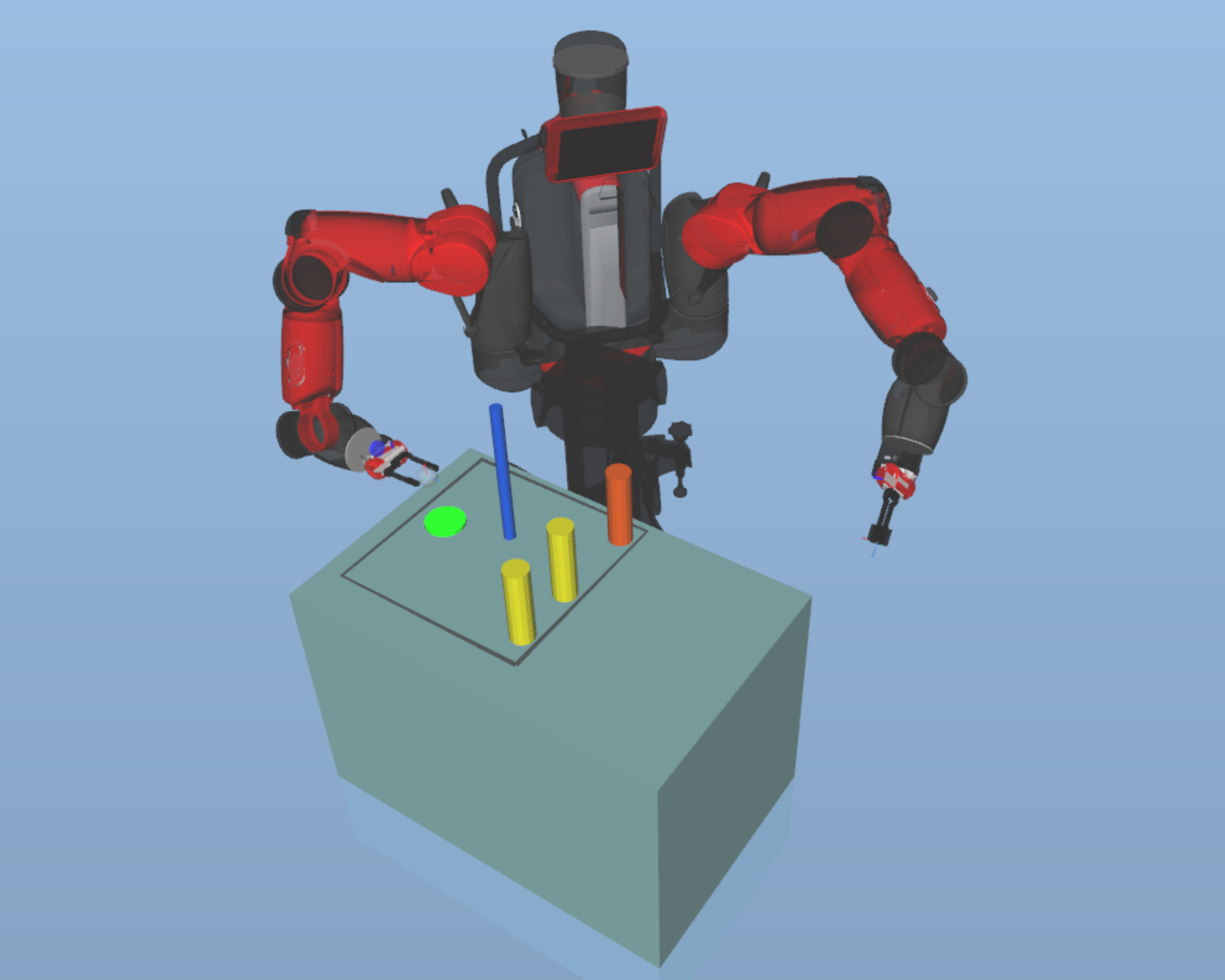}
  \includegraphics[clip,trim={2cm 5cm 20.5cm 5cm},width=.475\linewidth]{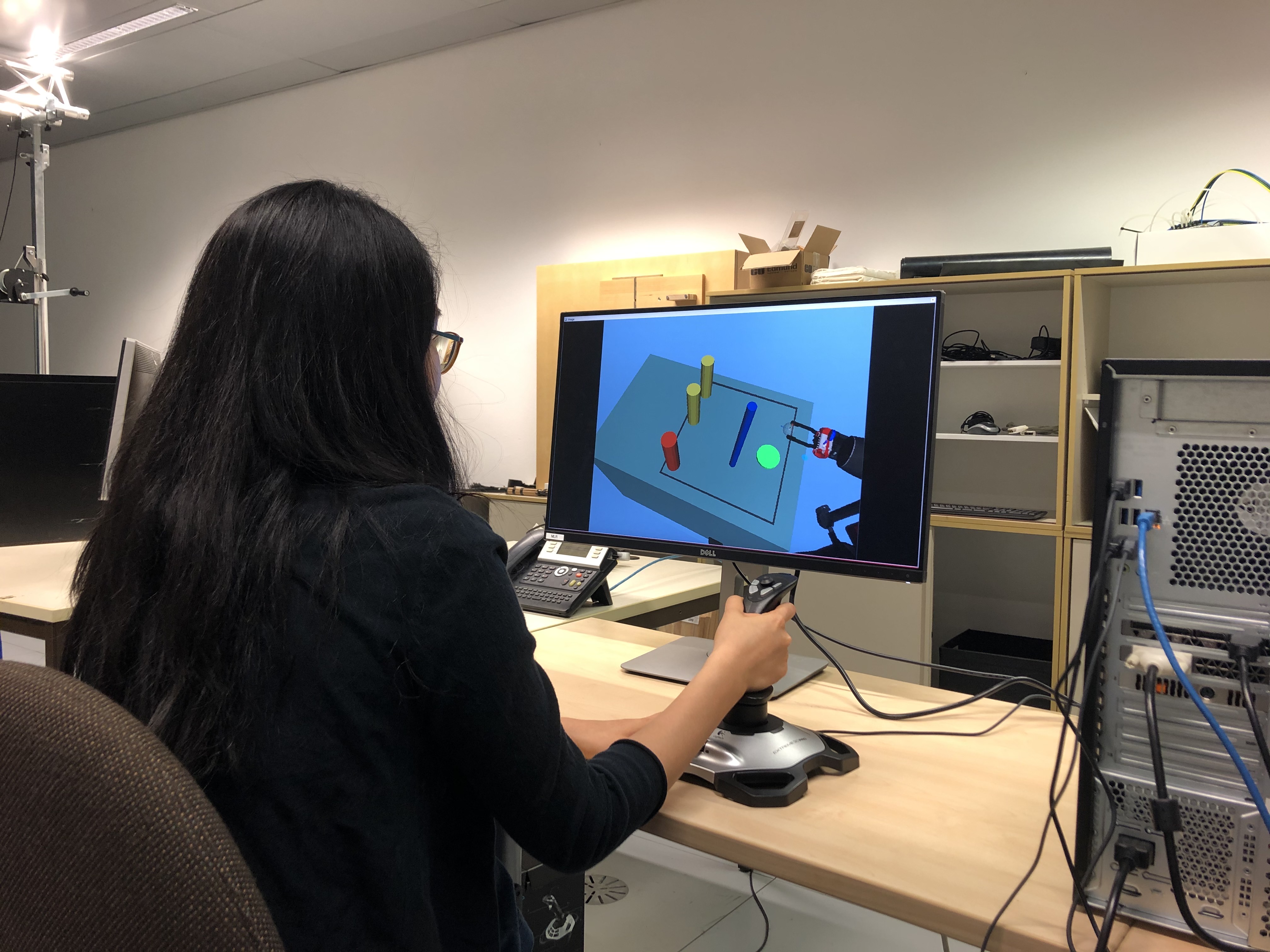}
  \caption{Teleoperation workspace setup (left) and user interaction using a joystick (right)}
  \label{fig:teleop_setup}
  \vspace{-.5cm}
\end{figure}

We perform a human participant study to assess the efficacy of our method. 
We define a simulated teleoperation environment, which consists of the Baxter robot that operates in a $50\text{cm}\times50\text{cm}$ workspace over a table (see Figure~\ref{fig:teleop_setup}) and multiple cylinders that represent objects and an obstacle. The user controls the robot's right arm end-effector using a joystick (Logitech Extreme 3D Pro). The robot is controlled at around 30Hz. Physical collisions are not simulated in the environment and grasping is simulated by attaching the object to the gripper when grasping is initiated. 
We hypothesize the following:
\begin{itemize}
  \item The natural gradient shared control method allows the user to take more control during the task while ensuring safety and efficiency.
\end{itemize}
 
\subsection{User Study Procedure}
The user study comprised of 16 participants (12 male, 4 female) with the right hand all being their dominant hand. The participants had no noticeable prior experience with robots and gave their informed consent before starting the study.

Users were asked to teleoperate the robot's gripper using the joystick to pick the red cylinder and bring it back to the green goal position while avoiding the blue pole as shown in~Figure~\ref{fig:teleop_setup}. The users controlled the robot's arm in task space: the gripper's velocity using pitch and roll motions and the z-axis rotation using the yaw motion of the joystick. Grasping was initiated using the trigger button. Users were provided with a perspective view of the robot's workspace (see Figure~\ref{fig:teleop_setup} right) to simulate a teleoperation scenario with limited camera view. 

We chose a within-subjects design, where each user participated in all systems. Each user performed three sets of demonstrations. Each set consisted of four different environments repeated over the control methods: in total 16 of episodes. The random order of teleoperation was predefined and balanced over the study. The teleoperation methods were as following:
\begin{itemize}
\item \textbf{DC}: Direct Control
\item \textbf{NG}: Natural Gradient Shared Control
\item \textbf{LB}: Linear Blending
\item \textbf{OA}: Obstacle Avoidance
\end{itemize}

For \textbf{LB}, we followed the ``timid" mode for linear arbitration suggested in Dragan et.al.~\cite{dragan2013policy}. The robot policy described in~\ref{ssec:learning_policy} was also  using the robot policy described in. 
Minimal assistance was provided to avoid the obstacle in
\textbf{OA} using a signed distance function as shown in Figure~\ref{fig:F_inv_ellipse}(c). 
In all cases, the speed of the action was determined by the user.

\section{RESULTS}
\label{sec:results}
\begin{figure}[h]   
  \captionsetup[subfigure]{justification=centering}
  \centering
  \subfloat[Duration]{\includegraphics[clip,trim={0.cm 2.5cm 0.5cm 1.41cm},width=.49\linewidth]{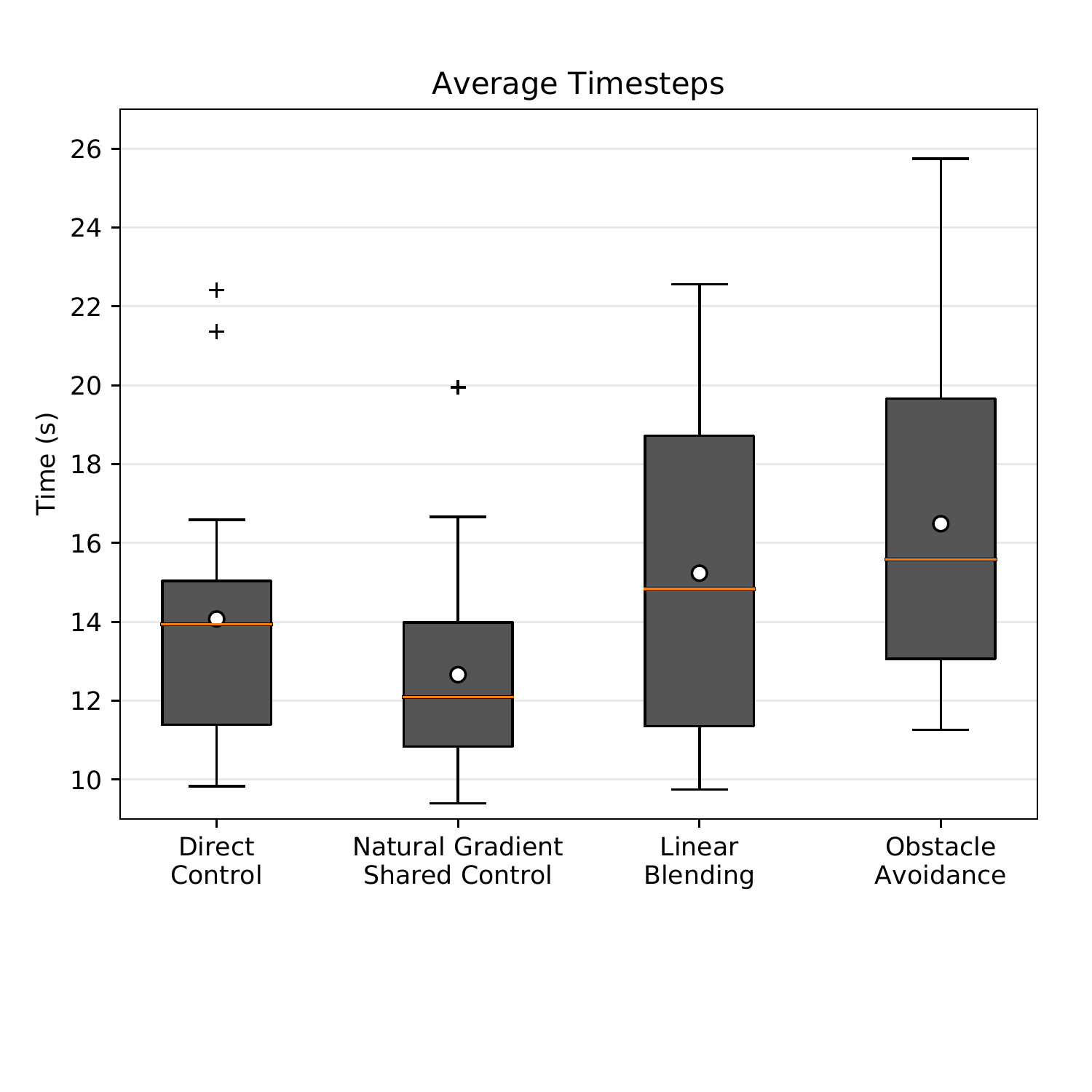}}
  \vspace{-.15cm}
  \centering
  \subfloat[Travel distance]{\includegraphics[clip,trim={0cm 2.5cm 0.5cm 1.41cm},width=.49\linewidth]{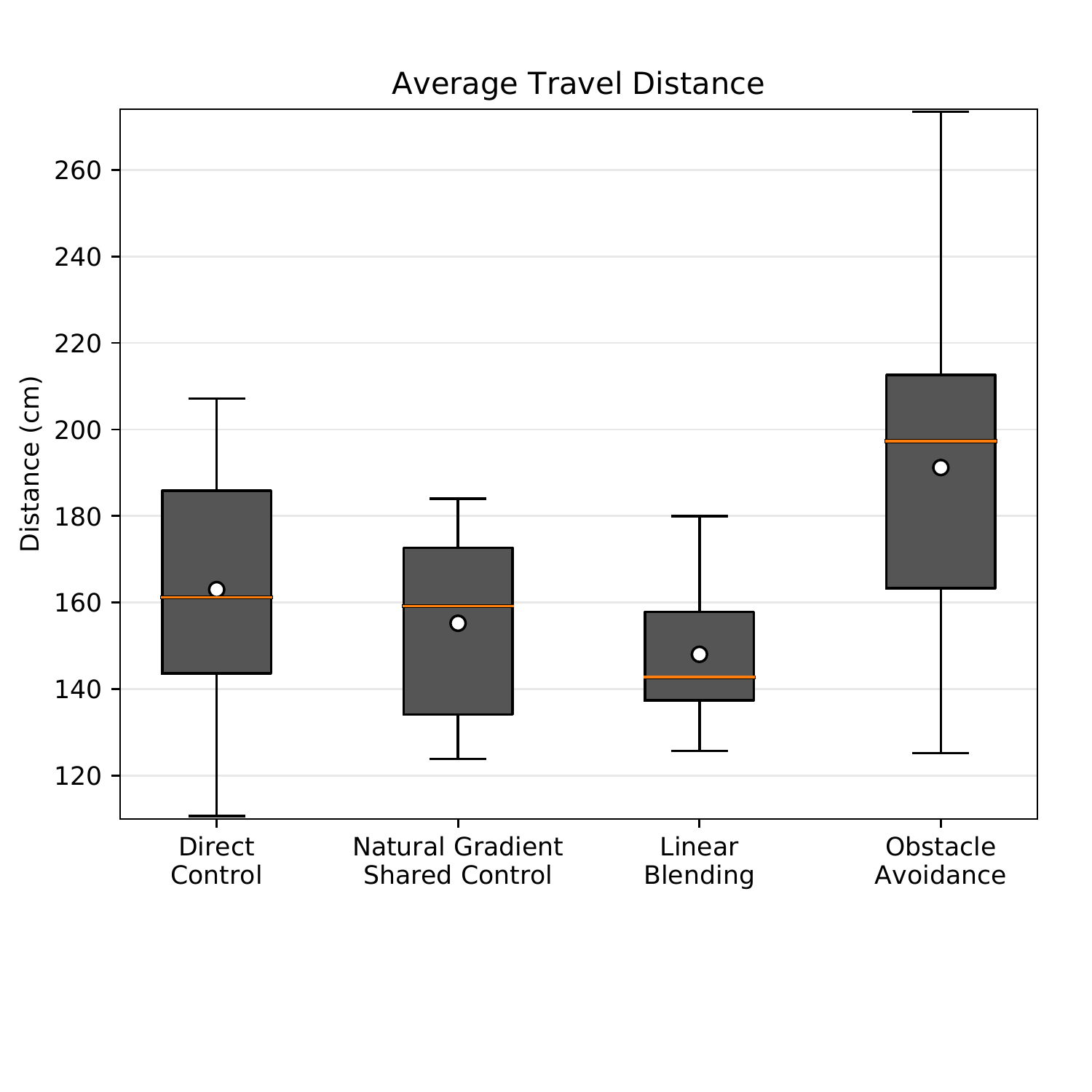}}\\
  \vspace{-.15cm}
  \centering
  \subfloat[Minimum proximity to obstacle]{\includegraphics[clip,trim={0cm 2.5cm 0.5cm 1.41cm},width=.49\linewidth]{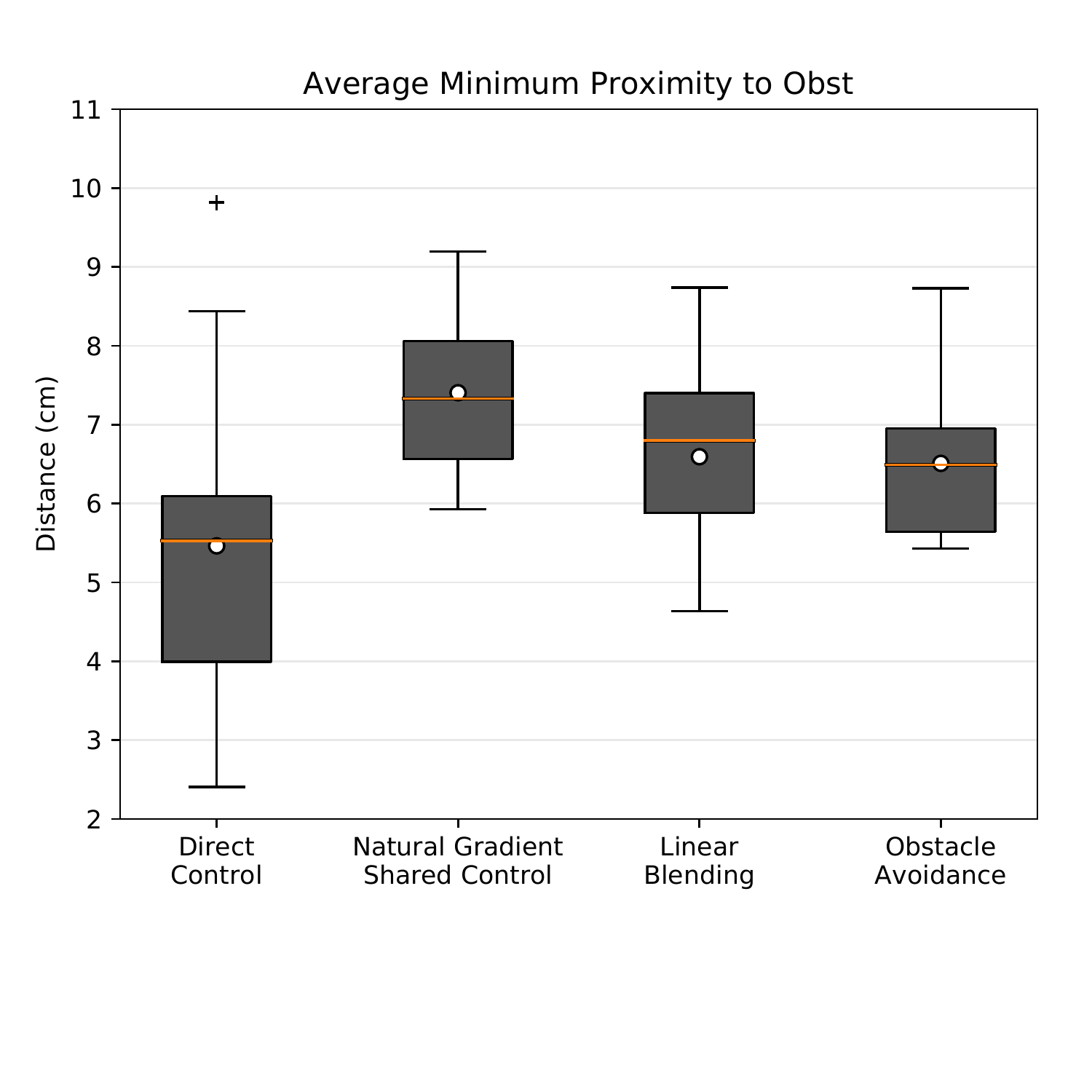}}
  \vspace{-.05cm}
  \centering
  \subfloat[Cosine distance]{\includegraphics[clip,trim={0cm 2.5cm 0.5cm 1.41cm},width=.49\linewidth]{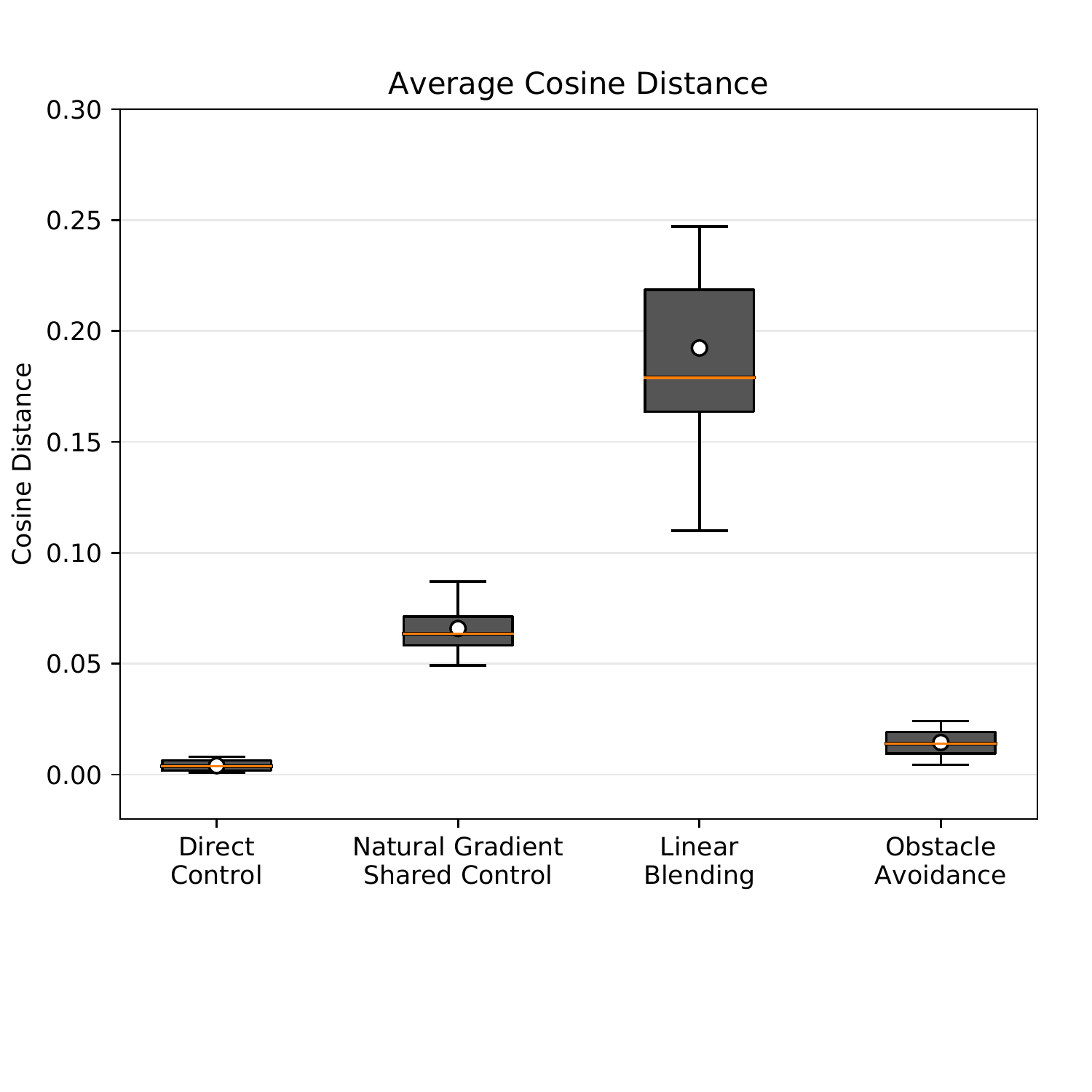}}
  \vspace{-.05cm}
  \caption{Comparison of control paradigms across all users for (a) execution time, (b) travel distance, (c) minimum proximity to obstacle, (d) cosine distance.}
  \label{fig:boxplots}
  \vspace{-0.5cm}
\end{figure}

We evaluate the results using a repeated measures analysis of variance (ANOVA), where the assistance method was used as a factor. We compared four quantitative measures as shown in Table~\ref{tab:table1} and Figure~\ref{fig:boxplots}: task duration, travel distance, minimum proximity to the obstacle, and the cosine distance between actions. The results show that our method was effective in assisting while allowing more user control authority. 
\begin{table}[h]
  \begin{center}
    \caption{User study results: Mean and standard deviation}
    \label{tab:table1}
    \begin{tabular}{c|c|c|c|c}
      \toprule 
      \textbf{Method} & \textbf{Duration} & \textbf{Travel Dist.} & \textbf{Proximity} & \textbf{Cosine Dist.} \\
      & (s) & (cm) & (cm) & ($\times$ \num{e-2}) \\
      \midrule 
      \textbf{DC} & 14.1 $\pm$ 3.5 & 163.0 $\pm$ 26.2 & 5.5 $\pm$ 1.8 & 0.4 $\pm$ 0.25\\
      \textbf{NG} & 12.7 $\pm$ 2.8 & 155.2 $\pm$ 20.5 & 7.4 $\pm$ 1.0 & 6.6 $\pm$ 1.0\\
      \textbf{LB} & 15.2 $\pm$ 4.0 & 148.0 $\pm$ 14.6 & 6.6 $\pm$ 1.1 & 19.2 $\pm$ 5.2\\
      \textbf{OA} & 16.5 $\pm$ 4.2 & 191.2 $\pm$ 40.4 & 6.5 $\pm$ 0.9 & 1.5 $\pm$ 0.6\\
      \bottomrule 
    \end{tabular}
  \end{center}
  \vspace{-1cm}
\end{table}

\begin{figure}[t] 
  \centering
  \captionsetup[subfigure]{justification=centering}
  \setbox1=\hbox{\includegraphics{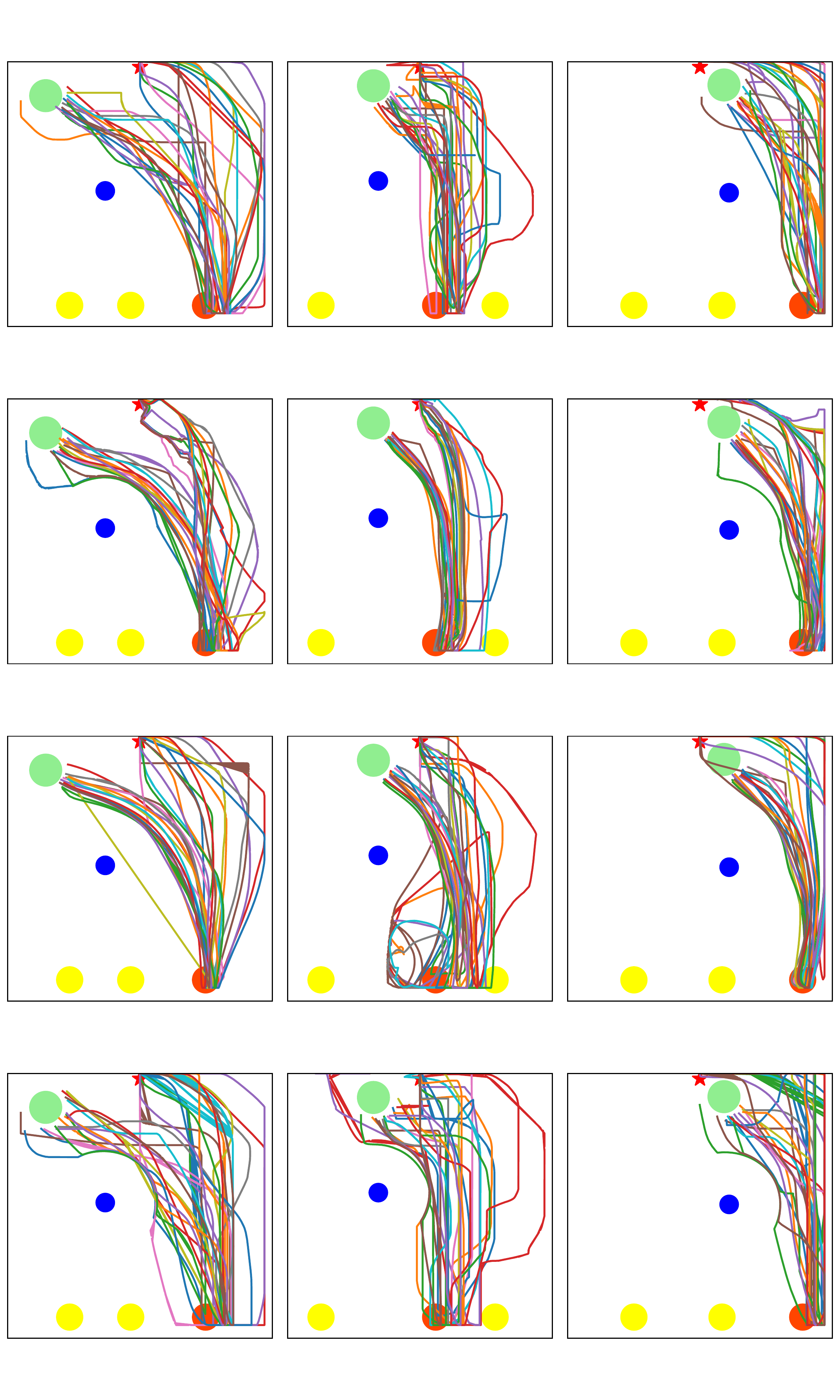}}%
  \subfloat[Direct control]{\centering\includegraphics[scale=0.295,viewport=0 .765\ht1 \wd1 0.97\ht1,clip]{env_8_1_2}}%
  \quad  
  \subfloat[Natural Gradient Shared Control]{\centering\includegraphics[scale=0.295,viewport=0 .525\ht1 \wd1 .73\ht1,clip]{env_8_1_2}}%
  \quad  
  \subfloat[Linear Blending]{\centering\includegraphics[scale=0.295,viewport=0 .285\ht1 \wd1 .49\ht1,clip]{env_8_1_2}}%
  \quad  
  \subfloat[Obstacle Avoidance]{\centering\includegraphics[scale=0.295,viewport=0 0.044\ht1 \wd1 .25\ht1,clip]{env_8_1_2}}\quad  
  \caption{Top-down visualization of user demonstrations for teloperation method (rows) of three environments (columns).
%
}
  \label{fig:traj_combined}
  \vspace{-.5cm}
\end{figure}

When comparing the average task duration in Figure~\ref{fig:boxplots}~(a), \textbf{NG} reported the shortest duration over all other methods with statistical significance: \textbf{DC}($f(4,12)=20.67, p=0.0004$), \textbf{LB}($f(4,12)=16.93, p=0.0009$), and \textbf{LB}($f(4,12)=61.50, p<0.0001$). When comparing the travel distance of the end-effector in Figure~\ref{fig:boxplots}~(b), \textbf{LB} showed the lowest average distance followed by \textbf{NG}. However, the results were statistically equivalent compared to \textbf{NG} ($f(4,12)=1.7933, p=0.2005$). This implies that \textbf{NG} is capable of providing assistance towards achieving efficient teleperation.

We compared the average minimum proximity between the gripper and the obstacle as a notion of safety. The closer distance denotes that it is more likely to collide to the obstacle. As shown in Figure~\ref{fig:boxplots}~(c), \textbf{NG} showed to provide highest obstacle avoidance compared to all other methods with statistical significance: \textbf{DC}($f(4,12)=31.42, p<0.0000$), \textbf{LB}($f(4,12)=18.7363, p=0.0006$), and \textbf{OA}($f(4,12)=23.2720, p=0.0002$).

The cosine distance represents the amount of disagreement between the user and the executed command. As shown in Figure~\ref{fig:boxplots}~(d), the average cosine distance of \textbf{NG} is significantly smaller than \textbf{LB} ($f(4,12)=104.36, p<0.0000$) suggesting that the method performed actions closer to what the user intended thus supporting our hypothesis that \textbf{NG} would allocate more control authority. \textbf{DC} and \textbf{OA} naturally showed the lowest in cosine distance since there was zero or minimum assistance.

In \textbf{LB}, assistance is provided such that the robot proceeds towards the optimal path. When the prediction is wrong with high confidence, the user must give an exaggerated opposing command to fight off the autonomous robot actions. This explains the results shown in Figure~\ref{fig:boxplots}, in which the travel distance was low but still showed large variance in task duration and cosine distance. 


Figure~\ref{fig:traj_combined} shows trajectories of user demonstrations in three environments. As seen from quantitative results, \textbf{DC} and \textbf{OA} trajectories exhibit divergent paths compared to \textbf{NG} or \textbf{LB} methods. Trajectories include dragging along the workspace boundary or approaching the place position from various directions. One notable characteristic of the \textbf{LB} method is shown in the second column of Figure~\ref{fig:traj_combined}. Since the target object and another object were relatively close in this environment, assistance was often provided towards the wrong goal. The user had to fight against the assistance and noisy trajectories are are near the objects.

Overall, \textbf{NG} showed reliable performance in task execution while still maintaining compliance with user commands. The results show that our method can be an option towards reducing the discrepancy and increasing user satisfaction during teleoperation.

\section{CONCLUSIONS}
\label{sec:conclusions}
We proposed a formalism for shared control that is based on
natural gradients emerging from the divergence constraint
between the user and robot policy. We introduced a method to approximate the Fisher information matrix, as a second derivative of the Q-function.
The efficacy of the system was demonstrated through a pilot user study and the initial results to quantitative measures showed that our approach allows efficient task completion while allowing more control authority to the user. We believe the results are convincing towards positive user experience and we plan to investigate it in our future work.

\addtolength{\textheight}{-1cm}   


\section*{ACKNOWLEDGMENT}

This work is partially funded by the research alliance ``System Mensch''.
The authors thank the International Max Planck Research School for Intelligent Systems (IMPRS-IS) for supporting Yoojin Oh.


\bibliographystyle{IEEEtran}
\bibliography{IEEEabrv,bibliography}{}

\begin{thebibliography}{10}
\providecommand{\url}[1]{#1}
\csname url@rmstyle\endcsname
\providecommand{\newblock}{\relax}
\providecommand{\bibinfo}[2]{#2}
\providecommand\BIBentrySTDinterwordspacing{\spaceskip=0pt\relax}
\providecommand\BIBentryALTinterwordstretchfactor{4}
\providecommand\BIBentryALTinterwordspacing{\spaceskip=\fontdimen2\font plus
\BIBentryALTinterwordstretchfactor\fontdimen3\font minus
  \fontdimen4\font\relax}
\providecommand\BIBforeignlanguage[2]{{%
\expandafter\ifx\csname l@#1\endcsname\relax
\typeout{** WARNING: IEEEtran.bst: No hyphenation pattern has been}%
\typeout{** loaded for the language `#1'. Using the pattern for}%
\typeout{** the default language instead.}%
\else
\language=\csname l@#1\endcsname
\fi
#2}}

\bibitem{dragan2013policy}
A.~D. Dragan and S.~S. Srinivasa, ``A policy-blending formalism for shared
  control,'' \emph{The International Journal of Robotics Research}, vol.~32,
  no.~7, pp. 790--805, 2013.

\bibitem{goil2013using}
A.~Goil, M.~Derry, and B.~D. Argall, ``Using machine learning to blend human
  and robot controls for assisted wheelchair navigation,'' in \emph{2013 IEEE
  13th International Conference on Rehabilitation Robotics (ICORR)}.\hskip 1em
  plus 0.5em minus 0.4em\relax IEEE, 2013, pp. 1--6.

\bibitem{anderson2014experimental}
S.~J. Anderson, J.~M. Walker, and K.~Iagnemma, ``Experimental performance
  analysis of a homotopy-based shared autonomy framework,'' \emph{IEEE
  Transactions on Human-Machine Systems}, vol.~44, no.~2, pp. 190--199, 2014.

\bibitem{gao2014contextual}
M.~Gao, J.~Oberl{\"a}nder, \emph{et~al.}, ``Contextual task-aware shared
  autonomy for assistive mobile robot teleoperation,'' in \emph{2014 IEEE/RSJ
  International Conference on Intelligent Robots and Systems}.\hskip 1em plus
  0.5em minus 0.4em\relax IEEE, 2014, pp. 3311--3318.

\bibitem{kim2011autonomy}
D.-J. Kim, R.~Hazlett-Knudsen, \emph{et~al.}, ``How autonomy impacts
  performance and satisfaction: Results from a study with spinal cord injured
  subjects using an assistive robot,'' \emph{IEEE Transactions on Systems, Man,
  and Cybernetics-Part A: Systems and Humans}, vol.~42, no.~1, pp. 2--14, 2011.

\bibitem{Javdani:2018bt}
S.~Javdani, H.~Admoni, \emph{et~al.}, ``Shared autonomy via hindsight
  optimization for teleoperation and teaming,'' \emph{The International Journal
  of Robotics Research}, vol.~37, no.~7, pp. 717--742, 2018.

\bibitem{broad2018operation}
A.~Broad, T.~Murphey, and B.~Argall, ``Operation and imitation under
  safety-aware shared control,'' in \emph{Workshop on the Algorithmic
  Foundations of Robotics}, 2018.

\bibitem{broad2019highly}
------, ``Highly parallelized data-driven mpc for minimal intervention shared
  control,'' in \emph{Robotics: science and systems}, 2019.

\bibitem{flemisch2016shared}
F.~Flemisch, D.~Abbink, \emph{et~al.}, ``Shared control is the sharp end of
  cooperation: Towards a common framework of joint action, shared control and
  human machine cooperation,'' \emph{IFAC-PapersOnLine}, vol.~49, no.~19, pp.
  72--77, 2016.

\bibitem{kofman2005teleoperation}
J.~Kofman, X.~Wu, \emph{et~al.}, ``Teleoperation of a robot manipulator using a
  vision-based human-robot interface,'' \emph{IEEE transactions on industrial
  electronics}, vol.~52, no.~5, pp. 1206--1219, 2005.

\bibitem{smith2008teleoperation}
C.~Smith, M.~Bratt, and H.~I. Christensen, ``Teleoperation for a ball-catching
  task with significant dynamics,'' \emph{Neural Networks}, vol.~21, no.~4, pp.
  604--620, 2008.

\bibitem{phillips2016autonomy}
C.~Phillips-Grafflin, H.~B. Suay, \emph{et~al.}, ``From autonomy to cooperative
  traded control of humanoid manipulation tasks with unreliable
  communication,'' \emph{Journal of Intelligent \& Robotic Systems}, vol.~82,
  no. 3-4, pp. 341--361, 2016.

\bibitem{muelling2015autonomy}
K.~Muelling, A.~Venkatraman, \emph{et~al.}, ``Autonomy infused teleoperation
  with application to bci manipulation,'' \emph{arXiv preprint
  arXiv:1503.05451}, 2015.

\bibitem{gopinath2016human}
D.~Gopinath, S.~Jain, and B.~D. Argall, ``Human-in-the-loop optimization of
  shared autonomy in assistive robotics,'' \emph{IEEE Robotics and Automation
  Letters}, vol.~2, no.~1, pp. 247--254, 2016.

\bibitem{jain2016approach}
S.~Jain and B.~Argall, ``An approach for online user customization of shared
  autonomy for intelligent assistive devices,'' in \emph{Proc. of the IEEE Int.
  Conf. on Robot. and Autom., Stockholm, Sweden}, 2016.

\bibitem{oh2019learning}
Y.~Oh, M.~Toussaint, and J.~Mainprice, ``Learning arbitration for shared
  autonomy by hindsight data aggregation,'' in \emph{Workshop on AI and Its
  Alternatives for Shared Autonomy in Assistive and Collaborative Robotics},
  2019.

\bibitem{reddy2018shared}
S.~Reddy, A.~D. Dragan, and S.~Levine, ``Shared autonomy via deep reinforcement
  learning,'' \emph{arXiv preprint arXiv:1802.01744}, 2018.

\bibitem{amari1998natural}
S.-I. Amari, ``Natural gradient works efficiently in learning,'' \emph{Neural
  computation}, vol.~10, no.~2, pp. 251--276, 1998.

\bibitem{amari1998why}
S.-I. Amari and S.~C. Douglas, ``Why natural gradient?'' in \emph{Proceedings
  of the 1998 IEEE International Conference on Acoustics, Speech and Signal
  Processing, ICASSP'98 (Cat. No. 98CH36181)}, vol.~2.\hskip 1em plus 0.5em
  minus 0.4em\relax IEEE, 1998, pp. 1213--1216.

\bibitem{kakade2002natural}
S.~M. Kakade, ``A natural policy gradient,'' in \emph{Advances in neural
  information processing systems}, 2002, pp. 1531--1538.

\bibitem{peters2008natural}
J.~Peters and S.~Schaal, ``Natural actor-critic,'' \emph{Neurocomputing},
  vol.~71, no. 7-9, pp. 1180--1190, 2008.

\bibitem{schulman2015trust}
J.~Schulman, S.~Levine, \emph{et~al.}, ``Trust region policy optimization,'' in
  \emph{International conference on machine learning}, 2015, pp. 1889--1897.

\bibitem{ratliff2013information}
\BIBentryALTinterwordspacing
N.~Ratliff, ``Information geometry and natural gradients,'' 2013. [Online].
  Available:
  \url{\url{https://ipvs.informatik.uni-stuttgart.de/mlr/wp-content/uploads/2015/01/}\newline\url{mathematics{\_}for{\_}intelligent{\_}}systems{\_}lecture12{\_}notes{\_}I.pdf}
\BIBentrySTDinterwordspacing

\bibitem{ly2017tutorial}
A.~Ly, M.~Marsman, \emph{et~al.}, ``A tutorial on fisher information,''
  \emph{Journal of Mathematical Psychology}, vol.~80, pp. 40--55, 2017.

\bibitem{Kristiadi2020blog}
\BIBentryALTinterwordspacing
A.~Kristiadi. (2018, Mar.) Natural gradient descent. [Online]. Available:
  \url{https://wiseodd.github.io/techblog/2018/03/14/natural-gradient/}
\BIBentrySTDinterwordspacing

\bibitem{martens2014new}
J.~Martens, ``New insights and perspectives on the natural gradient method,''
  \emph{arXiv preprint arXiv:1412.1193}, 2014.

\bibitem{ziebart2009planning}
B.~D. Ziebart, N.~Ratliff, \emph{et~al.}, ``Planning-based prediction for
  pedestrians,'' in \emph{2009 IEEE/RSJ International Conference on Intelligent
  Robots and Systems}.\hskip 1em plus 0.5em minus 0.4em\relax IEEE, 2009, pp.
  3931--3936.

\bibitem{mainprice2016warping}
J.~Mainprice, N.~Ratliff, and S.~Schaal, ``Warping the workspace geometry with
  electric potentials for motion optimization of manipulation tasks,'' in
  \emph{2016 IEEE/RSJ International Conference on Intelligent Robots and
  Systems (IROS)}.\hskip 1em plus 0.5em minus 0.4em\relax IEEE, 2016, pp.
  3156--3163.

\bibitem{kitani2012activity}
K.~M. Kitani, B.~D. Ziebart, \emph{et~al.}, ``Activity forecasting,'' in
  \emph{European Conference on Computer Vision}.\hskip 1em plus 0.5em minus
  0.4em\relax Springer, 2012, pp. 201--214.

\bibitem{atkeson1997locally}
C.~G. Atkeson, A.~W. Moore, and S.~Schaal, ``Locally weighted learning,'' in
  \emph{Lazy learning}.\hskip 1em plus 0.5em minus 0.4em\relax Springer, 1997,
  pp. 11--73.

\end{thebibliography}

\end{document}